\pdfoutput=1
\documentclass[11pt]{article}

\usepackage[preprint]{acl}

\usepackage{times}
\usepackage{latexsym}
\usepackage{multirow}

\usepackage[T1]{fontenc}
\usepackage[utf8]{inputenc}

\usepackage{microtype}

\usepackage{inconsolata}

\usepackage{graphicx}
\usepackage{amsmath}
\usepackage{booktabs}
\usepackage{algorithm}
\usepackage{algpseudocode}
\newtheorem{prop}{Proposition}
\newtheorem{duplicate}{Proposition}
\newcommand\blfootnote[1]{%
  \begingroup
  \renewcommand\thefootnote{}\footnote{#1}%
  \addtocounter{footnote}{-1}%
  \endgroup
}

\title{Cracking the Code of Hallucination in LVLMs with Vision-aware Head Divergence}

\author{
 \textbf{Jinghan He\textsuperscript{1,2}},
 \textbf{Kuan Zhu\textsuperscript{1,2}},
 \textbf{Haiyun Guo\textsuperscript{1,2}$^\ast$},
 \textbf{Junfeng Fang\textsuperscript{3}},
 \textbf{Zhenglin Hua\textsuperscript{4}},
 \\
 \textbf{Yuheng Jia\textsuperscript{4}},
 \textbf{Ming Tang\textsuperscript{1}},
 \textbf{Tat-Seng Chua\textsuperscript{3}},
 \textbf{Jinqiao Wang\textsuperscript{1,2,5}$^\ast$}
\\
 \textsuperscript{1}Foundation Model Research Center, Institute of Automation, Chinese Academy of Sciences
 \\
 \textsuperscript{2}School of Artificial Intelligence, University of Chinese Academy of Sciences
 \\
 \textsuperscript{3}National University of Singapore
 \textsuperscript{4}Southeast University
 \textsuperscript{5}Wuhan AI Research
\\
 \small{
   \texttt{hejinghan2022@ia.ac.cn}, \texttt{\{kuan.zhu, haiyun.guo, jqwang\}@nlpr.ia.ac.cn}
 }
}

\begin{document}
\maketitle
\blfootnote{$^\ast$ Corresponding author.}
\begin{abstract}
Large vision-language models (LVLMs) have made substantial progress in integrating large language models (LLMs) with visual inputs, enabling advanced multimodal reasoning. Despite their success, a persistent challenge is hallucination—where generated text fails to accurately reflect visual content—undermining both accuracy and reliability. Existing methods focus on alignment training or decoding refinements but primarily address symptoms at the generation stage without probing the underlying causes. In this work, we investigate the internal mechanisms driving hallucination in LVLMs, with an emphasis on the multi-head attention module. Specifically, we introduce Vision-aware Head Divergence (VHD), a metric that quantifies the sensitivity of attention head outputs to visual context. Based on this, our findings reveal the presence of vision-aware attention heads that are more attuned to visual information; however, the model's overreliance on its prior language patterns is closely related to hallucinations. Building on these insights, we propose Vision-aware Head Reinforcement (VHR), a training-free approach to mitigate hallucination by enhancing the role of vision-aware attention heads. Extensive experiments demonstrate that our method achieves superior performance compared to state-of-the-art approaches in mitigating hallucinations, while maintaining high efficiency with negligible additional time overhead. The code is available at \url{https://github.com/jinghan1he/VHR}.
\end{abstract}
\section{Introduction}

Large vision-language models (LVLMs) \cite{dai2023instructblip,liu2024improved} represent a notable advancement in artificial intelligence by enabling large language models (LLMs) to understand visual inputs. However, LVLMs still face the challenge of hallucination \cite{rohrbach2018object}, where generated text does not accurately correspond to visual content. This misalignment can compromise the accuracy and reliability of LVLMs across a wide range of vision and language tasks, limiting their practical applications \cite{you2024v2x}.

\begin{figure}[t]
\centering
\includegraphics[width=\linewidth]{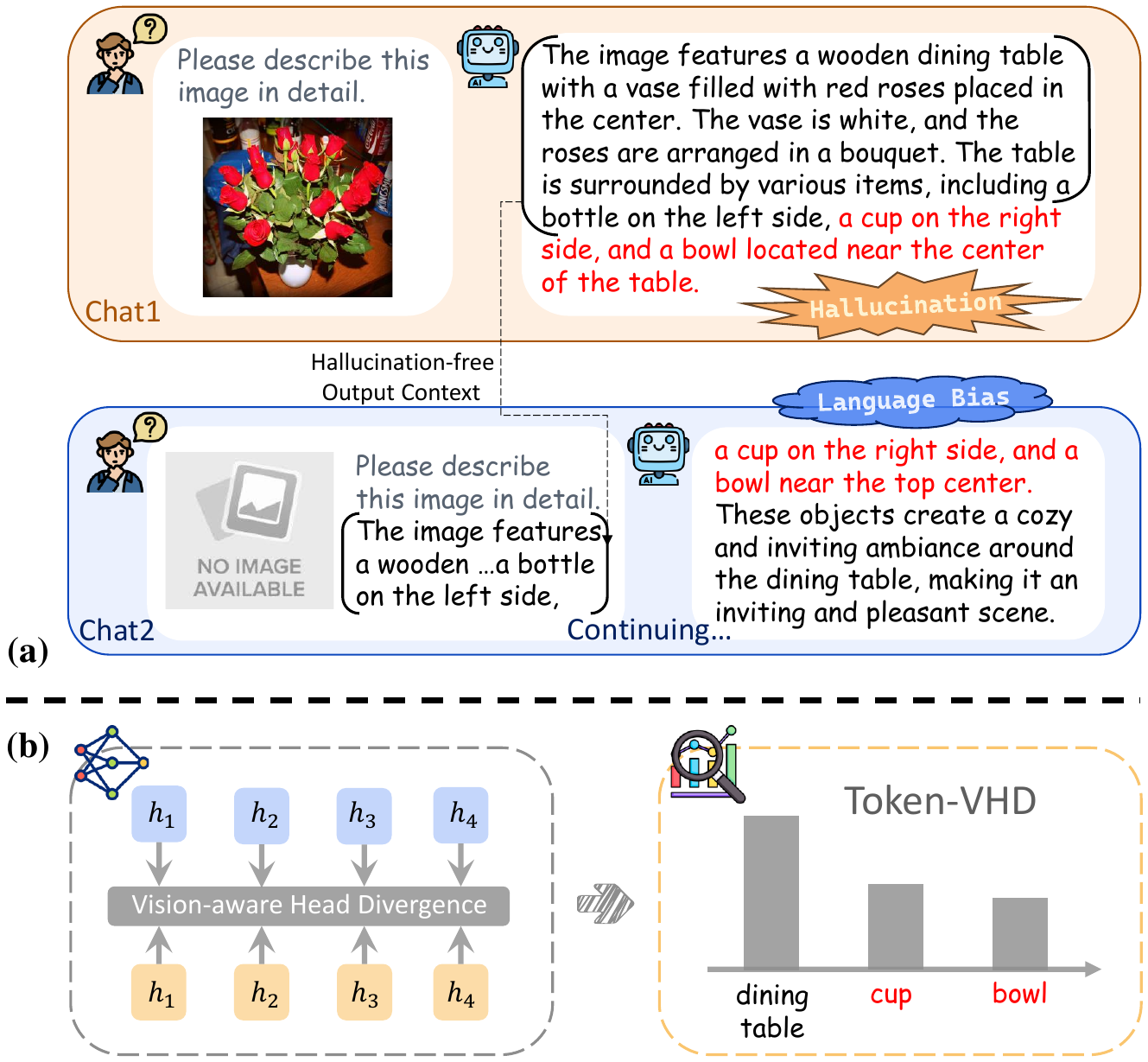}
\caption[]{(a) An example indicating the connection between hallucination in LVLMs and language bias. When hallucination occurs (chat 1), we remove the image input and prompt the model to complete the description (chat 2). The output closely resembles the hallucinated content\footnotemark. (b) The proposed VHD metric measures the sensitivity of the attention head outputs to image inputs, reflecting the degree of language bias. Hallucinated words generally correspond to lower T-VHD scores.}
\label{fig:motivation}
\end{figure}
\footnotetext{Details on this example are exhibited in Appendix \ref{sec:lb_exapmle}.}
To mitigate this issue, several approaches incorporate additional information or models for alignment training \cite{zhao2023beyond,yu2024rlhf} or post-processing \cite{zhou2023analyzing,yin2023woodpecker}, which incur higher training or inference costs. Recently, another line of research focuses on refining decoding strategies, employing methods like contrastive decoding \cite{leng2024mitigating,zhu2024ibd,kim2024code,gong2024damro} or beam search \cite{huang2024opera} to adjust the logits distribution during inference. However, these approaches merely intervene at the output level to rectify hallucinations after they occur, without directly targeting and adjusting the internal mechanisms that drive hallucinations. This work aims to fill this research gap.

One of the factors contributing to hallucination in LVLMs is their tendency to prioritize language patterns~\cite{ghosh2024visual,parcalabescu2024vision}, which can lead to the generation of fluent but inaccurate content. We further investigate this phenomenon and present an example in Figure \ref{fig:motivation}. Specifically, when prompted to continue generating an image description, the model generates highly consistent outputs, irrespective of whether an image is provided. This problem may arise from biased language patterns in the training data, which are incorporated into the model's parameters \cite{liu2024survey}, causing output to rely more on internal knowledge than image context. \citealp{yu2023characterizing} analyzed similar biases in language models and revealed that the multi-head attention module contains both in-context and memory heads. Manipulating these heads can influence whether the output is driven by contextual information or internal knowledge.

Building on previous findings, we are inspired to investigate the relationship between hallucination in LVLMs and the multi-head attention mechanism. To this end, we introduce a novel metric, Vision-aware Head Divergence (VHD), to quantify how the output of each attention head changes when the image context is removed in a generation step of LVLMs. Our analysis reveals that only a few heads show significant sensitivity to the image context, while the majority exhibit minimal variation. Based on this, we aggregate the VHD values from the most prominent attention heads in a generation step, resulting in the Token-VHD (T-VHD) metric. This metric allows us to evaluate the model's reliance on visual content versus language priors when predicting each token, as illustrated in the bottom-right part of Figure \ref{fig:motivation}. By examining the T-VHD scores, we observe that words and sentences associated with hallucinations generally correspond to lower values, further supporting the role of language bias in hallucination in LVLMs.

Leveraging the insights above, we propose Vision-aware Head Reinforcement (VHR), a training-free approach aimed at enhancing the model's reliance on visual context rather than language priors. This method proactively mitigates hallucination in LVLMs by first identifying key attention heads based on their VHD scores and then amplifying their contributions during generation. Theoretical analysis demonstrates that this scaling-up operation effectively re-orients the output of the multi-head attention module towards the reinforced head component, improving the alignment of the model's output with visual context. Experiments on established LVLM hallucination benchmarks show that VHR outperforms existing decoding strategies, validating its effectiveness and efficiency in alleviating hallucinations.

Our main contributions can be summarized as follows: 

\begin{itemize}
    \item We propose the VHD metric to probe the attention heads in LVLMs for the language-bias tendency, and the T-VHD metric to analyze the relationship between language-biased generation and hallucination in LVLMs. 
    \item We propose VHR, a training-free method that proactively mitigates hallucinations by adaptively identifying and reinforcing key attention heads during generation.
    \item Extensive experiments demonstrate that VHR outperforms existing decoding methods on widely-adopted hallucination benchmarks with negligible additional time cost.
\end{itemize} 

\section{Preliminary}

\textbf{LVLM generation.} The LVLMs take both image and text as input. The image is encoded into vision tokens using an image encoder and projected to the text embedding space through a connector. These vision tokens $x_V$ are then combined with tokenized text input $x_T$ and passed into the LLM component for autoregressive generation:
\begin{equation}
\label{eq:generate}
y_t = \arg\max p_\theta(y_t|y_{<t}, x_V, x_T),
\end{equation}
\noindent where $y_{<t}$ and ${y_t}$ denote the earlier and the currently generated text tokens, respectively.

\textbf{Multi-head attention.} The multi-head attention mechanism is a core component of transformer models with each attention head performing the self-attention operation among tokens:
\begin{equation}
\label{eq:attention}
% \small
\resizebox{\linewidth}{!}{
$\begin{aligned}
    & A_{l,i}(X_{l,i}) = \text{Attention}(X_{l,i}W_{l,i}^Q, X_{l,i}W_{l,i}^K, X_{l,i}W_{l,i}^V), \\ 
    &\text{where} \, \text{Attention}(Q, K, V)=\text{softmax}(\frac{QK^T}{\sqrt{d_k}})V. \\
\end{aligned}$
}
\end{equation}
$X_{l,i}$ and $A_{l,i}$ represent the input and output of the $i$-th attention head in the $l$-th layer, respectively. $W^Q$, $W^K$, and $W^V$ denote the learned weight matrices for the query, key, and value transformations, respectively. $d_k$ is the dimension of the query ($Q$) and key ($K$) vectors. The outputs of all the attention heads in the $l$-th layer are then concatenated and linearly transformed into the output space of this module:
\begin{equation}
\label{eq:mha}
\resizebox{\linewidth}{!}{
    $\text{MHA}_l(X_l) = [A_{l,1}(X_{l,1}), \cdots, A_{l,n_h}(X_{l,n_h})]W_l^O,$
}
\end{equation}
where $n_h$ denotes the number of attention heads in each layer, $X_l$ is the input to the MHA module in the $l$-th layer, and $W^O$ is the learned weight matrices for the output linear transformation. 

\textbf{Attention head output during generation.} To more clearly delineate the correspondence between the model's intermediate outputs and its inputs, we combine Equation \ref{eq:generate} and \ref{eq:attention} to introduce the notation $A_{l,i}(y_t|y_{<t},x_V,x_T)$. This notation represents the output of the $i$-th head in the $l$-th layer for generation step $t$, given the inputs $x_V$ and $x_T$, and the generation history $y_{<t}$.

\begin{figure}[t]
\centering
  \includegraphics[width=\linewidth]{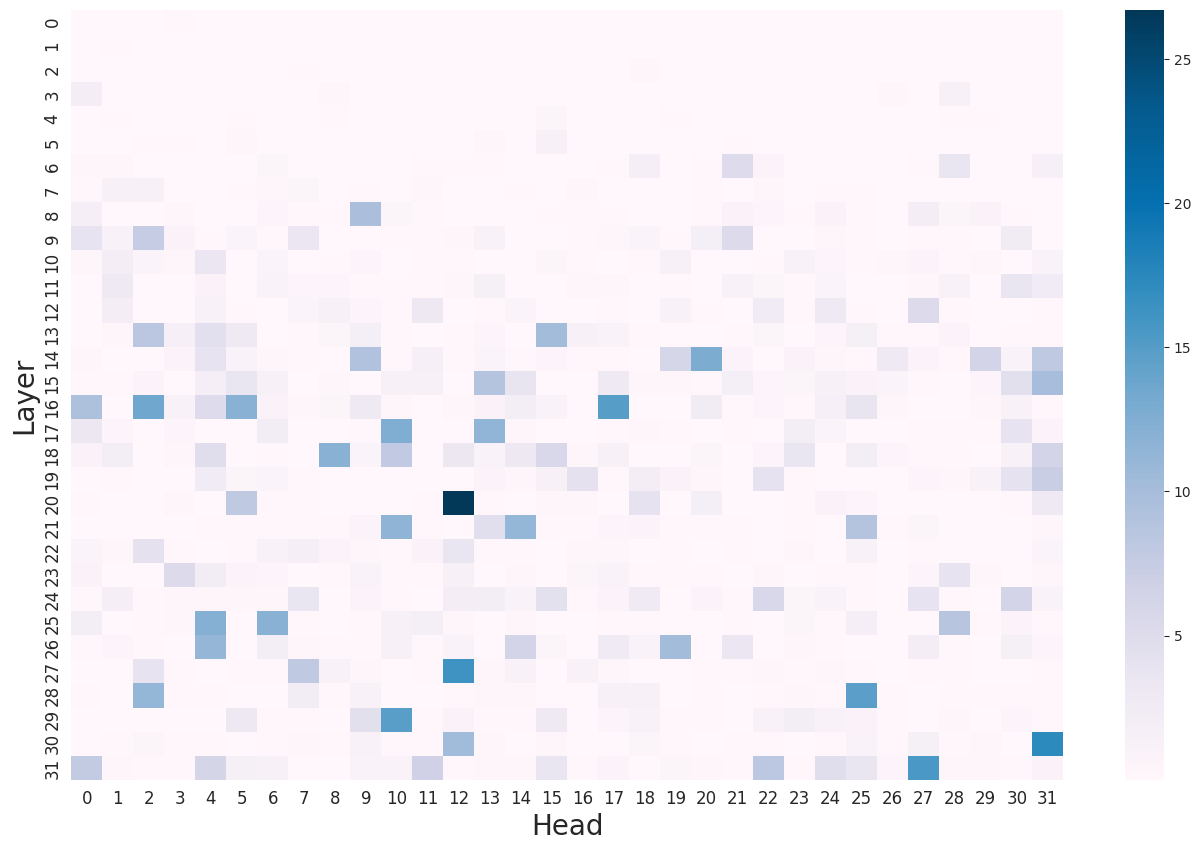}
  \caption{VHD scores of all the attention heads among all layers at one generation step. }
  \label{fig:layer_head}
\end{figure}

\section{Method}

\subsection{Vision-aware Head Identification}
\label{sec:vhd}
\textbf{Vision-aware head divergence (VHD).} Inspired by the presence of in-context and memory heads in the model \cite{yu2023characterizing}, we investigate whether different attention heads exhibit significantly different degrees of sensitivity to visual content. Specifically, we propose the vision-aware head divergence metric, which measures the change in the output of attention head for generation step $t$ when the image context is removed: 
\begin{equation}
\resizebox{\linewidth}{!}{
\label{eq:vhd}
    $\text{VHD}_{l,i} = \text{d}\left(A_{l,i}(y_t|y_{<t},x_V,x_T),\, A_{l,i}(y_t|y_{<t},x_T)\right),$
}
\end{equation}
where $d$ represents the Euclidean distance measure \cite{tabak2014geometry}. 

Figure \ref{fig:layer_head} visualizes the VHD scores for each attention head in the model. Specifically, we prompt LLaVA-1.5 with an image and the instruction "Please describe the image in detail" to generate descriptions, calculating the VHD scores when predicting the first token. The results show that a few attention heads exhibit notably higher VHD scores, while the others show minimal sensitivity. This suggests the presence of vision-aware attention heads that are more attuned to visual information. More examples of VHD scores during the generation process are presented in Appendix \ref{sec:vhd_examples}. 

\begin{figure}[t]
\centering
  \includegraphics[width=\linewidth]{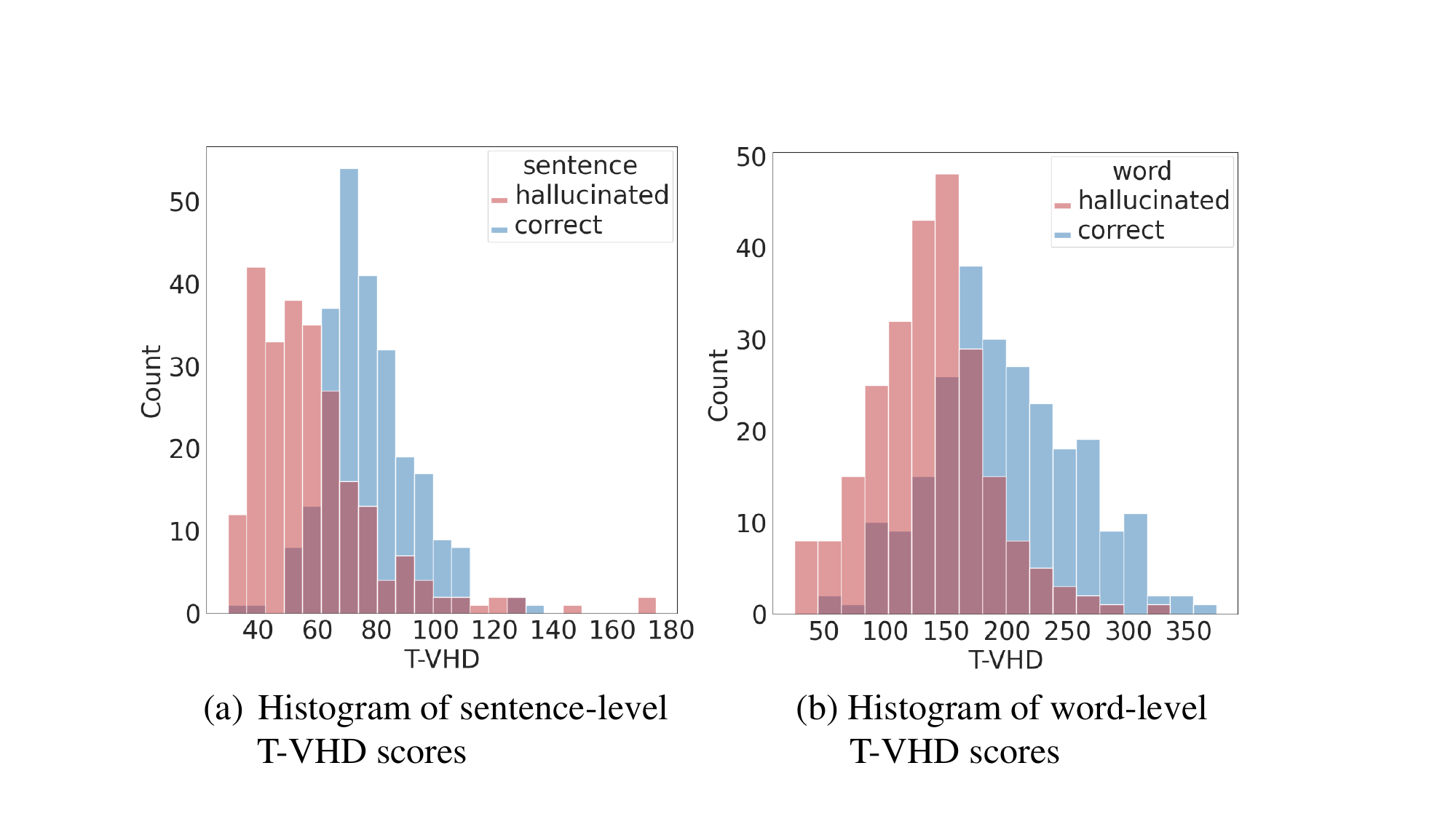}
  \caption{Relationship between T-VHD scores and hallucinations in LVLMs. Sentences and words associated with hallucinations generally correspond to lower T-VHD scores. Best viewed in color.}
  \label{fig:vhd_statistics}
\end{figure}

\begin{figure*}[t]
\centering
  \includegraphics[width=\linewidth]{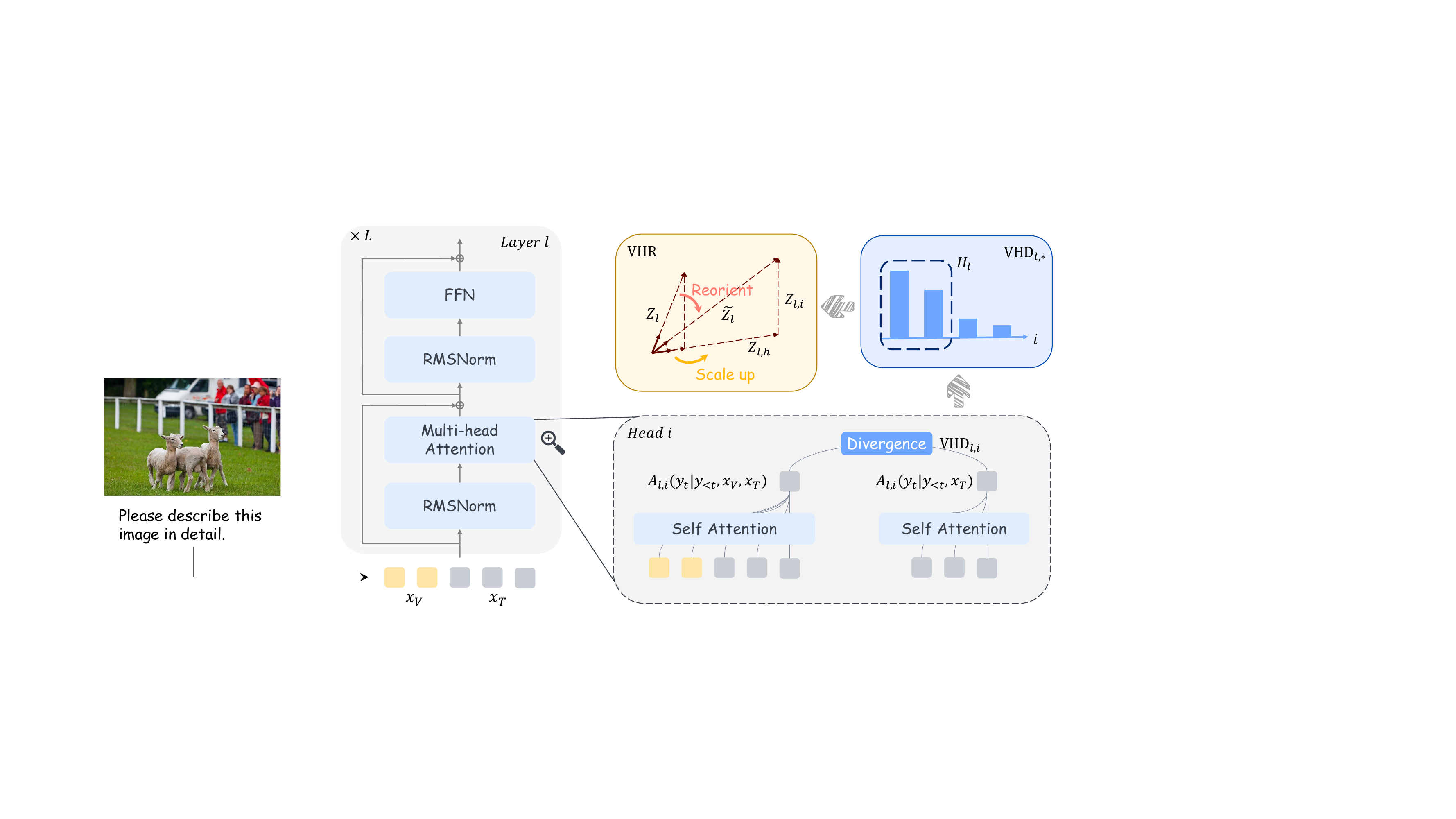}
  \caption{The illustration of the proposed VHD metric and the VHR approach to mitigate hallucinations in LVLM. We select the attention heads that are sensitive to visual information for a given layer based on the VHD metric, \textit{i.e.} $H_l$, and then amplify their outputs to reinforce their contributions.}
  \label{fig:framework}
\end{figure*}

\textbf{Token-VHD.} Beyond the varying degrees of vision awareness within the model, we further explore whether the VHD scores vary across different token generation steps. To this end, the VHD scores of the most prominent attention heads in each layer of the model are aggregated into the Token-VHD metric:
\begin{equation}
    \text{T-VHD} = \sum_{l}\sum_{i} \text{topk}_{i}(\text{VHD}_{l,i}, k).
\end{equation}
Note that we only consider the top $k$ VHD scores at each layer to ensure that the large number of insensitive attention heads does not dilute the aggregation metric. Eventually, T-VHD metric serves as an indicator of the model's reliance on visual information over language priors when predicting a specific token. 

Leveraging this metric, we can quantitatively analyze the relationship between hallucination in LVLMs and language bias at different levels of granularity, specifically at the sentence and word levels. To this end, we conduct an experiment on a random sample of 500 images from the CHAIR benchmark \cite{rohrbach2018object}, tracking the T-VHD scores at each generation step. Object-related words in the generated descriptions are classified as either hallucinated or correct, depending on whether they appear in the annotated object set for the given image. Sentences are then labeled according to whether they contain any hallucinated words. Figure \ref{fig:vhd_statistics} presents the experimental results, highlighting the distributional differences in T-VHD scores between hallucinated and correct instances. These findings provide statistical evidence that language bias is closely related to hallucinations in LVLMs.

\subsection{Vision-aware Head Reinforcement}

Since only a small subset of attention heads within the model are sensitive to visual information, we can amplify their contributions during generation to strengthen the model's reliance on visual cues and counteract language bias. 
As discussed in Section \ref{sec:vhd}, the VHD metric effectively captures the sensitivity of attention heads to visual information, making it a suitable indicator for selecting key attention heads for reinforcement. However, we observed that some high VHD values stem from a surge in the activation of attention heads upon the removal of visual context, indicating negative vision sensitivity. Amplifying the contributions of such heads would diverge from our objective. Therefore, we propose to zero out these undesired outliers, \textit{i.e.} $\text{VHD}_{l,i}=0$, if the following condition satisfies:
\begin{equation}
\begin{aligned}
\label{eq:outliers}
    &\begin{cases}
    \;\text{VHD}_{l,i}  > \mu(\text{VHD}_{l,*})+\sigma(\text{VHD}_{l,*}), \\
    \;\text{TA}_{l,i} >\mu(\text{TA}_{l,*})+\sigma(\text{TA}_{l,*}), \\
    \end{cases}\\
    &\quad\; \text{where} \;\text{TA}_{l,i}=\|A_{l,i}(y_t|y_{<t},x_T)\|^2.
\end{aligned}
\end{equation}
$\mu$ and $\sigma$ represent the mean and standard deviation. 
Next, for the multi-head attention module in a given layer of the model, we select the first half of the attention heads based on their VHD scores and directly scale up their outputs by a factor of $\alpha$:
\begin{equation}
\resizebox{0.89\linewidth}{!}{
$\begin{aligned}
\label{eq:vhr}
    &\widetilde{A}_{l,i} =
    \begin{cases}
    \alpha \cdot A_{l,i},\quad &\text{if}\; i \in H_l, \\
    A_{l,i}, &\text{otherwise}, \\
    \end{cases} \\
    & \text{where} \;H_l = \{i|\; \text{VHD}_{l,i}>\text{median}(\text{VHD}_{l,*})\}. \\
\end{aligned}$
}
\end{equation}

\textbf{Apply VHR layer by layer.} This specific implementation allows for the selection and reinforcement of attention heads within a single forward pass, as opposed to first selecting the heads in all layers and then reinforcing them in two separate passes. Additionally, when VHR is applied across multiple layers simultaneously, the reinforcement in earlier layers can influence the VHD scores of subsequent layers. The layer-by-layer VHR strategy helps to avoid such inconsistencies, as the previous layers are already reinforced when calculating the VHD scores for a given layer.

\textbf{Determine the heads at the first generation step.} Although we can compute the VHD scores and select the key heads at each generation step, reinforcing different heads at different steps may bring negative effects. Specifically, LVLMs rely on KV caching to speed up inference, which means that the keys and values of the previous tokens will not be recalculated in subsequent generation steps. Therefore, the important heads should be determined at the beginning of the generation process to ensure consistency in the $Q$, $K$, and $V$ of all tokens in the attention module. Our experimental results show that this approach is sufficient to mitigate hallucinations.

\textbf{Comparison with other head identification methods.} Different from existing attention head identification methods in the field of model interpretability \cite{yu2023characterizing,li2024inference,zhou2024role,fang2024towards,he2024seekr}, VHR does not require any annotation and can adaptively detect the key heads for each sample. Furthermore, rather than identifying and reinforcing all key attention heads in the model in two forward passes, VHR ensures computational efficiency and metric consistency by iteratively applying the \textit{select-then-reinforce} approach across the layers. The complete procedure of VHR is provided in Algorithm \ref{alg:vhr}.

\subsection{Attention Output Reorientation} 
\label{sec:reorient}
Scaling up the outputs of certain attention heads within a layer to reinforce its contribution is a straightforward and intuitive operation, and we present a theoretical analysis to substantiate its rationale. Consider the input to the FFN module following the MHA module in layer $l$, which can be expressed as follows:
\begin{equation}
\label{eq:ffn}
\begin{aligned}
    Z_l &=\text{RMSNorm}(\hat X_l+\text{MHA}_l(X_l)) \\
        &= \hat{g}_l\cdot \frac{\hat X_l+\text{MHA}_l(X_l)}{\|\hat X_l+\text{MHA}_l(X_l)\|},
\end{aligned}
\end{equation}
where $\hat g_l$ is a fixed constant after training, and $\hat X_l$ is the input to the $l$-th layer before RMSNorm. Due to the normalization operation, only the direction of the overall output from earlier modules is crucial. 
\begin{prop}
\label{prop:1}
Consider a layer $l$ within an LVLM, and let $h$ be the index of the attention head to be reinforced. Let $\widetilde{Z}_l$ be the input to the FFN module obtained with $\widetilde{A}_{l,h}=\alpha \cdot A_{l,h}$ 
 ($\alpha > 1$), $Z_l$ be the original input obtained with $A_{l,h}$, $Z_{l,h}$ be the pseudo-input obtained with only the $A_{l,h}$ component. Then it holds that $cos(\widetilde{Z}_l,Z_{l,h})>cos(Z_l, Z_{l,h})$. 
\end{prop}
The proof is detailed in Appendix \ref{sec:proof}. Proposition \ref{prop:1} implies that amplifying the output of a specific head in the MHA module effectively reorients the direction of $Z_l$ towards the output direction of the reinforced head component. This provides theoretical support for the mechanism underlying the reinforcement of the key attention head. An overview of the proposed VHD metric and the VHR method is presented in Figure \ref{fig:framework}.

\begin{algorithm}[t]
\caption{VHR}
 \textbf{Input} image $x_V$, instruction $x_T$, generation step $t$, scale factor $\alpha$, layers to reinforce $L_r$
\begin{algorithmic}[1]
    \For{layer $l \in L_r$}
        \If{$t=0$}
            \State Compute $\text{VHD}_{l,*}$ \Comment{Equation \ref{eq:vhd}}
            \State Zero out $\text{VHD}_{l,i}$ if Equation \ref{eq:outliers} holds
            \State Select the heads as $H_l$ \Comment{Equation \ref{eq:vhr}}
        \EndIf
        \State Reinforce the heads in $H_l$ \Comment{Equation \ref{eq:vhr}}
    \EndFor
\end{algorithmic}
\label{alg:vhr}
\end{algorithm}
\section{Experiments}

\begin{table*}[t]
\centering
\resizebox{\linewidth}{!}{
\begin{tabular}{l|ccc|ccc|ccc}
\toprule
 & \multicolumn{3}{c|}{InstructBLIP} & \multicolumn{3}{c|}{LLaVA-1.5} & \multicolumn{3}{c}{LLaVA-NeXT}  \\

 & CHAIR$_S\downarrow$& CHAIR$_I\downarrow$ & Len & CHAIR$_S\downarrow$ & CHAIR$_I\downarrow$ & Len & CHAIR$_S\downarrow$ & CHAIR$_I\downarrow$ & Len \\
\midrule
 Greedy & \underline{45.32}$_{\pm2.24}$ & \underline{12.98}$_{\pm0.76}$ & 91.06
        & 49.68$_{\pm1.47}$ & 14.32$_{\pm0.78}$ & 83.06
        & 29.08$_{\pm2.09}$ & 8.08$_{\pm0.74}$  & 157.06 \\
 Beam   & 48.56$_{\pm1.66}$ & 13.50$_{\pm0.44}$ & 94.87
        & 53.84$_{\pm2.41}$ & 15.60$_{\pm0.46}$ & 87.47
        & \underline{25.72}$_{\pm2.17}$ & 6.92$_{\pm0.88}$  & 160.64\\
 DoLa   & 46.00$_{\pm1.87}$ & 13.00$_{\pm0.91}$ & 90.75
        & 50.88$_{\pm2.34}$ & 14.64$_{\pm0.90}$ & 82.41
        & 28.76$_{\pm2.58}$ & 8.12$_{\pm0.78}$  & 155.75\\
 VCD    & 50.72$_{\pm2.44}$ & 14.42$_{\pm0.99}$ & 90.39
        & 51.92$_{\pm1.87}$ & 15.42$_{\pm0.84}$ & 83.12
        & 30.80$_{\pm2.48}$ & 8.72$_{\pm0.94}$  & 157.72\\
 OPERA  & 45.76$_{\pm2.32}$ & 13.06$_{\pm0.88}$ & 92.46
        & 44.28$_{\pm0.95}$ & 13.36$_{\pm0.47}$ & 75.88
        & - & - & - \\
 CODE   & 50.76$_{\pm2.06}$ & 14.12$_{\pm0.93}$ & 88.57
        & 47.96$_{\pm0.80}$ & 14.26$_{\pm0.57}$ & 78.52
        & 27.84$_{\pm2.73}$ & 7.98$_{\pm0.92}$ & 151.51\\
 EAH    & 46.40$_{\pm1.15}$ & 13.13$_{\pm0.60}$ & 92.33
        & \underline{38.76}$_{\pm2.47}$ & \underline{11.05}$_{\pm0.81}$ & 86.28
        & 28.13$_{\pm1.13}$ & \textbf{6.62}$_{\pm0.49}$  & 142.75 \\
 \midrule
 VHR    & \textbf{37.76}$_{\pm2.76}$ & \textbf{9.75}$_{\pm0.98}$ & 106.49
        & \textbf{33.32}$_{\pm1.31}$ & \textbf{9.71}$_{\pm0.45}$ & 81.33
        & \textbf{24.96}$_{\pm2.09}$ & \underline{6.80}$_{\pm0.59}$ & 156.92\\
\bottomrule
\end{tabular}
}
\caption{CHAIR evaluation results on MSCOCO dataset averaged over 5 random splits, with best in \textbf{bold} and second-best \underline{underlined}. \textit{Len} represents the average number of words in the generated descriptions.}
\label{tab:chair}
\end{table*}

\subsection{LVLMs}
We conduct experiments on three of the most representative LVLMs, \textit{i.e.} InstructBLIP-7b \cite{dai2023instructblip}, LLaVA-1.5-7b \cite{liu2024improved}, and LLaVA-NeXT-7b \cite{liu2024llavanext}. LVLMs are typically composed of an image encoder, a connector, and an LLM. Specifically, LLaVA-1.5-7b and LLaVA-NeXT-7b leverage MLP to align the visual and textual embedding space and feed all the image tokens from the image encoder to the LLM. In contrast, InstructBLIP uses Q-Former to reduce the number of image tokens before passing them to the LLM. LLaVA-NeXT differs from LLaVA-1.5 by offering a higher image resolution, allowing it to capture more visual details.

\subsection{Benchmarks}

\textbf{CHAIR.} The Caption Hallucination Assessment with Image Relevance (CHAIR) metric \cite{rohrbach2018object} evaluates object hallucination in image captioning by comparing generated captions with ground truth data. It identifies objects mentioned in captions but absent in images and calculates their proportion to quantify hallucination. Specifically, CHAIR includes two metrics at both caption level (CHAIR$_S$) and object level (CHAIR$_I$): 
\begin{equation}
\begin{aligned}
    \text{CHAIR}_S&=\frac{|\{\text{caption w/ hallucinated objects}\}|}{|\{\text{all captions}\}|}, \\
    \text{CHAIR}_I&=\frac{|\{\text{hallucinated objects}\}|}{|\{\text{all mentioned objects}\}|}.
\end{aligned}
\end{equation}
We randomly sample 500 images from the COCO 2014 validation set and repeat the experiments for five times with different random seeds. The LVLMs are prompted with "Please describe this image in detail." to get the descriptions. We report the average results for each metric along with the standard deviation.

\textbf{POPE.} POPE \cite{li2023evaluating} is a dataset for evaluating object hallucinations by having models answer true or false questions about the presence of objects in images. The dataset includes 500 images from MSCOCO \cite{lin2014microsoft}, with each image paired with questions like "Is there a <object> in the image?". The evaluation consists of three splits—random, popular, and adversarial—where objects are sampled in different ways. The evaluation metrics include Accuracy, Precision, Recall, and F1 scores, with the results averaged across all three splits. 

\textbf{LLaVA-Bench.} LLaVA-Bench (In-the-Wild) \cite{liu2024improved} is a comprehensive benchmark designed to evaluate the performance of vision-language models on a wide range of challenging tasks. It includes 24 images across diverse domains, such as indoor and outdoor scenes, memes, accompanied by 60 carefully crafted questions covering simple question answering, detailed descriptions, and complex reasoning. Due to the open-ended nature and complexity of the responses, we prompt the GPT-4V model to evaluate the LVLMs' outputs in terms of accuracy, detailedness, and naturalness. 

\subsection{Baselines}
We compare VHR with the popular training-free methods that do not introduce external information or models: DoLa \cite{chuang2023dola} derives the next-token distribution by contrasting the logits from later and earlier layers; VCD \cite{leng2024mitigating} contrasts the output distribution generated from the original and distorted image; OPERA \cite{huang2024opera} mitigates over-trust in previous summary tokens in beam-search decoding; CODE \cite{kim2024code} uses self-generated descriptions as contrast references to improve alignment with the actual visual content; EAH \cite{zhang2024seeing} enhances the attention sinks on image tokens in shallow layers. In addition, we also compare the performance of base LVLMs using greedy and beam search decoding. 

\subsection{Implementation Details}
We set $\alpha$ to 2 to strike a balance between effectively correcting hallucinations and minimizing the invasiveness of hidden states manipulation. VHR is applied to the second and last 14 layers for LLaVA series and the last 18 layers for InstructBLIP. We faithfully reproduced all baseline methods based on their open-source repositories and set the hyperparameters according to the values reported in the papers. The results of all methods are reported under consistent conditions of base models, prompts, and generation parameters to ensure a fair comparison. Specifically, the max\textunderscore new\textunderscore token is set to 512, and the number of beams is set to 5 for all methods involving beam search. 

\subsection{Results}

\textbf{CHAIR.} Table \ref{tab:chair} presents the performance of VHR in comparison to all baseline approaches on the CHAIR benchmark. The results for OPERA on LLaVA-NeXT are absent due to its excessive memory requirements. VHR demonstrates robust performance across all three LVLMs, achieving reductions of up to  16.36 in $\text{CHAIR}_S$ and 4.61 in $\text{CHAIR}_I$ on LLaVA-1.5. Notably, with increased image resolution and enhanced model capabilities, LLaVA-NeXT already exhibits a significant reduction in hallucinations compared to other base LVLMs, but VHR continues to exhibit notable effectiveness in mitigating its hallucinations. Moreover, VHR consistently outperforms baseline methods with greater stability, requiring only minor trade-offs in the length or richness of the generated description.

\textbf{POPE.}
As shown in Table \ref{tab:pope}, VHR outperforms all other decoding methods across all LVLMs. While the binary (yes/no) benchmark limits the full demonstration of VHR's strength in handling language bias, our method still consistently improves performance across models of varying capabilities. This indicates that VHR is a robust and effective training-free strategy for enhancing models at different performance levels.

\textbf{LLaVA-Bench.} 
The GPT-4V evaluation results on LLaVA-Bench (In-the-Wild) are presented in Table \ref{tab:llavawild}. These results demonstrate that VHR improves model accuracy on highly diverse and challenging tasks while preserving a consistent level of detailedness and naturalness. Additionally, the \textit{Len} metric in Table~\ref{tab:chair} and the qualitative results in Figure~\ref{fig:qualitative} also indicate that VHR has only a minor impact on language generation quality, despite its enhancement of vision-aware attention heads.

\begin{table}[t]
\centering
\resizebox{\linewidth}{!}{
\begin{tabular}{l|c|c|c}
\toprule
 & InstructBLIP & LLaVA-1.5 & LLaVA-NeXT  \\
\midrule
 Greedy & \underline{85.36} & 84.98 & \underline{88.51} \\
 Beam   & 84.40 & 85.30 & 87.97 \\
 DoLa   & 85.21 & 85.07 & 88.46 \\
 VCD    & 84.67 & 84.41 & 88.11 \\
 OPERA  & 84.41 & \underline{85.45} & - \\
 CODE   & 84.80 & 84.63 & 88.44  \\
 EAH    & 85.18 & 85.03 & 84.28\\
 \midrule
 VHR & \textbf{85.52} & \textbf{85.47} & \textbf{88.87}\\
\bottomrule
\end{tabular}
}
\caption{F1 scores on POPE averaged over popular, adversarial, and random splits, with best in \textbf{bold} and second-best \underline{underlined}.}
\label{tab:pope}
\end{table}

\begin{table}[t]
\centering
\resizebox{\linewidth}{!}{
\begin{tabular}{l|ccc}
\toprule
 & Accuracy & Detailedness & Naturalness \\
\midrule
InstructBLIP & 4.917 & 5.017 & 6.717\\
w/ VHR & \textbf{5.250} & \textbf{5.117} & \textbf{6.733}\\
\midrule
LLaVA-1.5 & 6.017 & 6.100 & \textbf{7.400}\\
w/ VHR & \textbf{6.333} & \textbf{6.217} & 7.333\\
\midrule
LLaVA-NeXT & 5.383 & \textbf{6.750} & \textbf{7.900} \\
w/ VHR & \textbf{5.783} & 6.700 & 7.833\\
\bottomrule
\end{tabular}
}
\caption{LLaVA-Bench (In-the-Wild) evaluation results, scored by GPT-4o via pairwise response comparison.}
\label{tab:llavawild}
\end{table}

\begin{table}[t]
\centering
\resizebox{\linewidth}{!}{
\begin{tabular}{l|l|cc}
\toprule
 Model & Method & CHAIR$_S \downarrow$ & CHAIR$_I \downarrow$ \\
\midrule
\multirow{3}{*}{InstructBLIP} & VHR & \textbf{37.76} & \textbf{9.75} \\
 & fixed VHR & 45.40 & 13.57 \\
 & outlier VHR & 37.76 & 10.18 \\
\midrule
\multirow{3}{*}{LLaVA-1.5} & VHR & \textbf{33.32} & \textbf{9.71} \\
 & fixed VHR & 44.72 & 13.81 \\
 & outlier VHR & 36.88 & 10.36 \\
\midrule
\multirow{3}{*}{LLaVA-NeXT} & VHR & \textbf{24.96} & \textbf{6.80} \\
 & fixed VHR & 36.96 & 9.80 \\
 & outlier VHR & 24.64 & 6.37 \\
\bottomrule
\end{tabular}
}
\caption{Ablation study on adaptively determining key heads per sample and removing outlier VHD scores.}
\label{tab:ablate-dynamic}
\end{table}

\begin{figure}[t]
\centering
  \includegraphics[width=\linewidth]{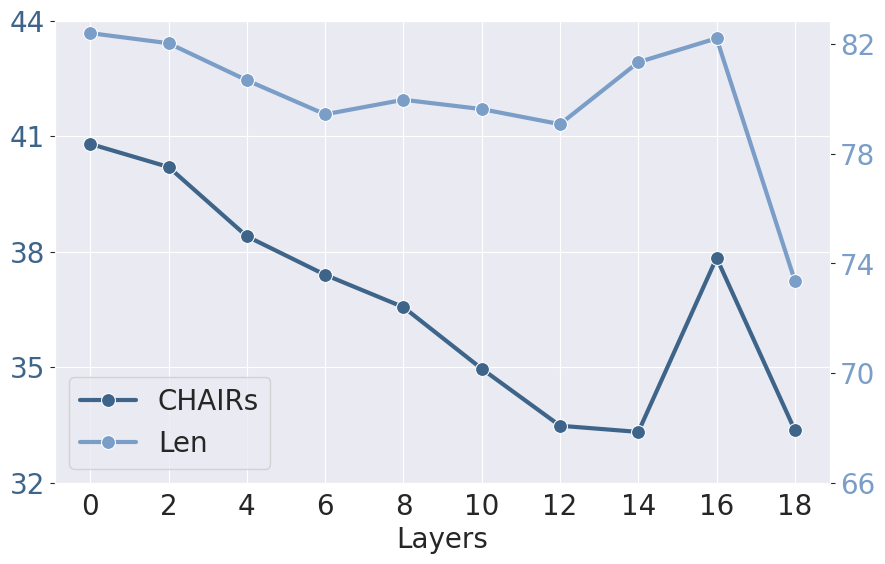}
  \caption{Results of VHR applied across different numbers of the last few layers in LLaVA-1.5.}
  \label{fig:layer_ablate}
\end{figure}

\subsection{Ablation Study}

\textbf{Impact of Adaptive Attention Head Selection.} Since VHR reinforces different attention heads for each sample, we conduct an ablation study to validate the necessity of this strategy. As shown in Table \ref{tab:ablate-dynamic}, fixing a set of attention heads identified by one sample for reinforcement across all samples leads to a significant performance drop. 

\textbf{Impact of Outlier VHD Score Removal.} VHR relies on VHD scores to identify critical attention heads, but these scores measure absolute influence, conflating both positive and negative vision sensitivity. To selectively enhance heads that improve visual context awareness, we propose excluding outlier VHD scores during selection (Equation~\ref{eq:outliers}). As shown in Table~\ref{tab:ablate-dynamic}, this refinement consistently matches or outperforms the baseline, demonstrating that pruning unhelpful sensitivity is essential for optimal head selection.

\textbf{Impact of Reinforced Layers.} Figure \ref{fig:layer_ablate} shows the ablation study results on the number of the last few layers for reinforcement. Increasing the number of reinforced layers continuously alleviates hallucinations, with optimal performance reached at the last 14 layers. However, further reinforcement degrades the model's generation quality and fails to effectively mitigate hallucinations. More discussion on the choice of reinforced layers and scale factor can be found in Appendix \ref{sec:layer_choice} and \ref{sec:scale_factor}.

\begin{figure}[t]
\centering
  \includegraphics[width=\linewidth]{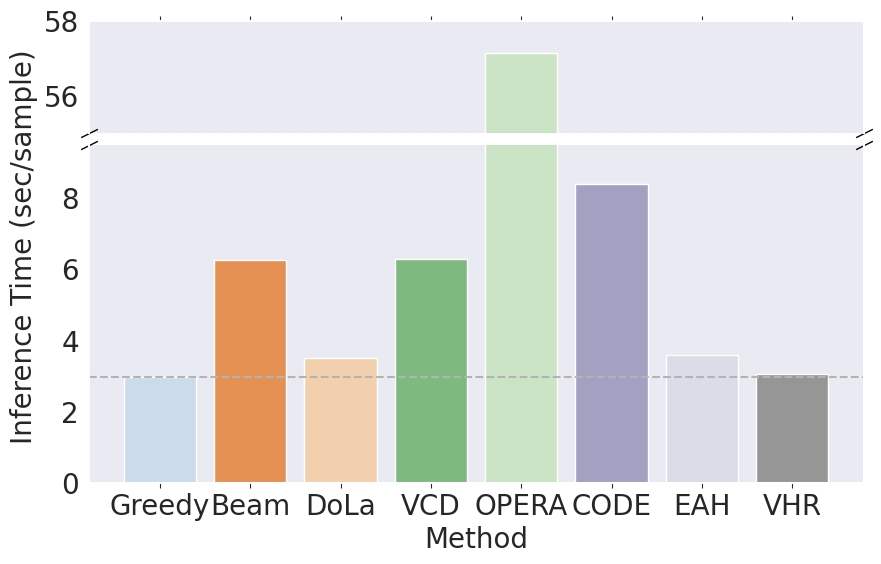}
  \caption{Comparison of inference time for different methods. }
  \label{fig:time_analysis}
\end{figure}

\subsection{Further Analysis} 
\textbf{Additional Time Analysis.}
For the first generation step, VHR requires an additional forward pass, removing the image context to calculate the VHD scores. In subsequent generation steps, only the scaling operation is needed. As a result, the extra computation introduced by VHR is negligible throughout the entire generation process. A detailed inference time comparison between VHR and baseline methods is presented in Figure \ref{fig:time_analysis}.

\begin{figure}[t]
\centering
  \includegraphics[width=\linewidth]{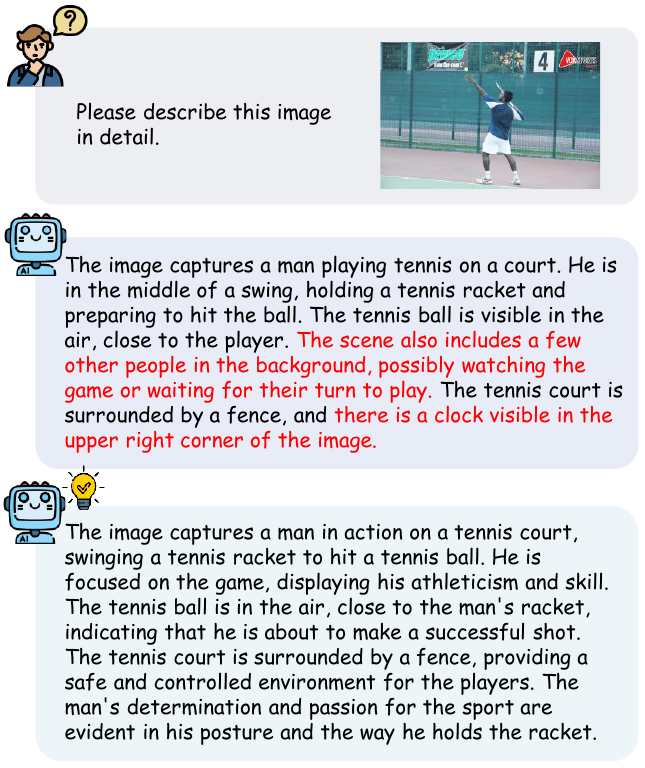}
  \caption{An example of VHR eliminating the hallucinated content.}
  \label{fig:qualitative}
\end{figure}

\textbf{Qualitative Results.}
To clearly demonstrate the effect of VHR in reducing hallucinations, we provide a concrete example in Figure \ref{fig:qualitative}. Without VHR, the LVLM generates content that is absent from the image, such as mentioning people watching the game in the background. This could stem from inherent language bias in the training data. When VHR is applied, the outputs are more accurately aligned with the actual content of the image.

\section{Related Work}

\subsection{Hallucination in LVLMs}

Hallucination in LVLMs refers to discrepancies between the generated text and the actual content present in the corresponding image. This issue stems from multiple factors, including data bias, insufficient image grounding by vision encoders, and misalignment between modalities~\cite{liu2024survey}. Current approaches to mitigating hallucinations in LVLMs can be broadly classified into three categories: training alignment \cite{zhao2023beyond,yu2024rlhf}, post-processing \cite{zhou2023analyzing,yin2023woodpecker,fang2024alphaedit}, and decoding strategies \cite{leng2024mitigating,zhu2024ibd,huang2024opera,kim2024code,gong2024damro}. The first two categories often rely on external resources or models to improve performance, whereas our proposed VHR method operates solely based on the model's internal states, without requiring additional external information. In contrast to existing decoding strategies, VHR intervenes proactively within the model to address hallucinations before they occur, rather than modifying the logits distribution at the output stage. A recent approach, EAH~\cite{zhang2024seeing}, also targets the attention mechanism to reduce hallucinations, but it focuses specifically on the attention sink phenomenon. In comparison, VHR reduces language bias in LVLMs by exploiting cross-modal contrast and re-adjusting the contributions of attention heads across multiple layers. Consequently, the underlying motivations and methodologies of the two works differ significantly.

\subsection{Language Bias in LVLMs}

Language bias refers to the tendency of models to prioritize language patterns or prior knowledge over the actual visual context presented in the input. This issue has a long-standing research history predating the emergence of LVLMs and was studied through methods like balanced multimodal training~\cite{goyal2017making} and causal inference~\cite{niu2021counterfactual}. With the advent of LVLMs, language bias has become an even more pressing concern. These models are typically pre-trained on massive corpora of text data, further exacerbating the potential for language overfitting. Studies have identified that LVLMs, despite their capabilities in visual recognition, struggle to fully integrate visual context~\cite{parcalabescu2024vision}, leading to hallucinations, particularly in reasoning tasks~\cite{ghosh2024visual}. To address this growing challenge, several recent methods have proposed solutions such as contrastive decoding~\cite{leng2024mitigating,zhu2024ibd} and visual description grounding decoding~\cite{ghosh2024visual}. However, these approaches directly manipulate the output logits, which introduces instability during generation and lacks a thorough analysis of the model's internal mechanisms. In contrast, our approach intervenes directly within the model, providing a more interpretable and effective way to address the internal factors driving language bias in LVLMs while complementing existing methods.

\section{Conclusion}

This work investigates the connection between hallucination in LVLMs and the multi-head attention mechanism. We introduce the VHD metric, which quantifies the sensitivity of attention head outputs to visual context, revealing that language bias can contribute to hallucinations in LVLMs. Building on these findings, we propose VHR, a training-free approach that strengthens the role of vision-aware attention heads to mitigate hallucinations. Extensive experiments demonstrate that VHR outperforms existing methods, significantly improving the alignment of LVLMs with visual information. 

\section*{Limitations}

Our analysis and mitigation strategy primarily focus on the multi-head attention mechanism of LVLMs. While this is a critical component influencing hallucinations, there may be other architectural factors—such as those in the vision encoder and the FFN module in the LLMs—that contribute to hallucinations but were not directly addressed in this study. Future work could focus on more comprehensive interventions that span the entire model, going beyond attention head manipulation.

\section*{Acknowledgements}

This work was supported by National Key R\&D Program of China under Grant No.2024YFC3015501, also partly supported by National Natural Science Foundation of China under Grant No.62276260, Beijing Municipal Science and Technology Project under Grant Z231100007423004, Beijing Natural Science Foundation under Grant 4244099, Aeronautical Science Foundation of China under Grant 2024M0710M0002.

\bibliography{custom}

\appendix

\section{A Case Study of Language Bias}

\begin{figure}[t]
\centering
  \includegraphics[width=\linewidth]{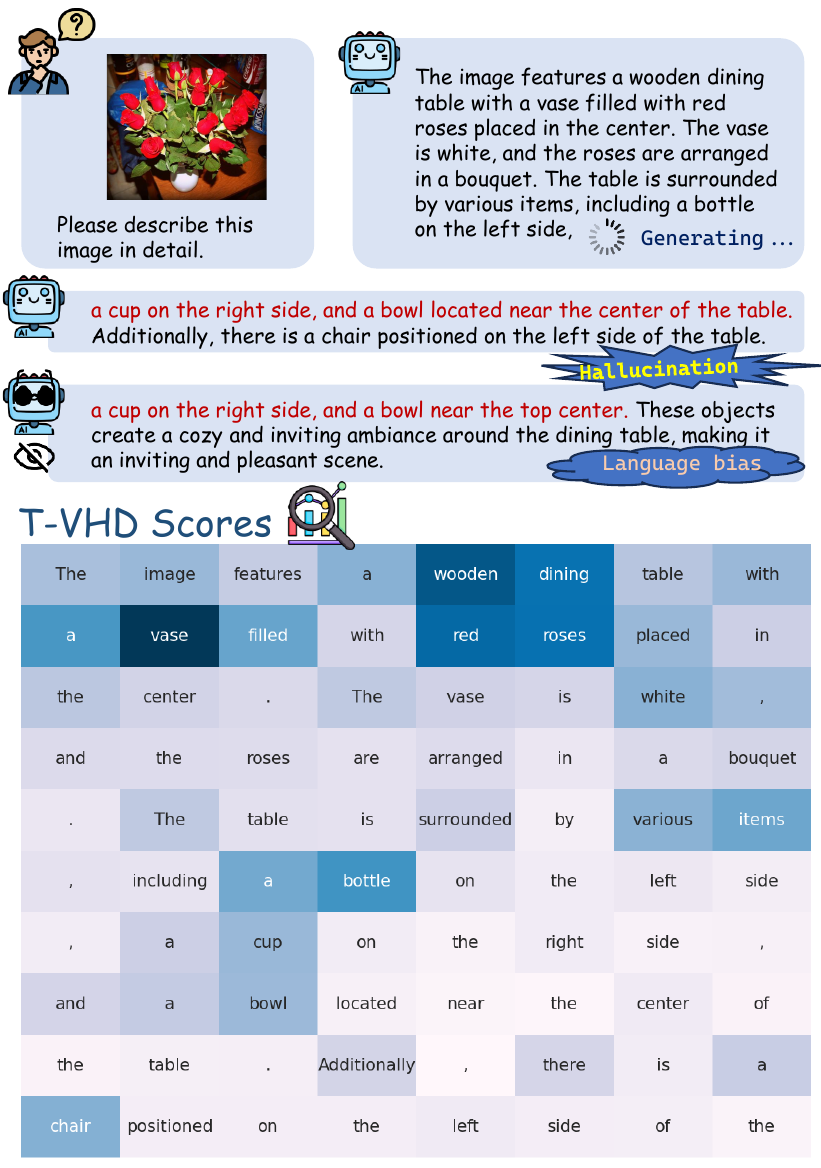}
  \caption{An example indicating the connection between hallucination in LVLMs and language bias. The proposed T-VHD metric reflects the model's reliance on visual content versus language priors at the token level.}
  \label{fig:detail_example}
\end{figure}

Figure \ref{fig:detail_example} illustrates an example that reflects language bias in LVLMs. We first prompted the model to describe the image, resulting in a complete description. We observed that the latter part of this description contained hallucinated content. To investigate the role of language bias in this case, we removed the image input and re-prompted the model using only the original text prompt and the non-hallucinated part of the initial response. This allowed us to observe how the model would continue the text based solely on its internal knowledge. We found that the continuation closely resembled the previously hallucinated content, suggesting it was generated based on internal language priors rather than visual evidence. This example clearly demonstrates the connection between language bias and hallucination in LVLMs. 

In the lower part of Figure \ref{fig:detail_example}, we visualize the T-VHD scores for each word in the generated description. The word color intensity reflects the T-VHD scores, with darker shades indicating higher sensitivity to visual input. The results show that words within fixed phrases (e.g., \textit{surrounded by}, \textit{on the right/left side}) generally have lower T-VHD scores, suggesting stronger reliance on language priors. In contrast, object terms (e.g. \textit{wooden dining table}, \textit{vase}, \textit{red roses}) typically exhibit higher T-VHD scores upon first mention, indicating greater dependence on visual context. Notably, among all object terms, hallucinated items (e.g., \textit{cup}, \textit{bowl}, \textit{chair}) tend to have lower T-VHD scores, highlighting reduced visual grounding. This example demonstrates how the proposed T-VHD metric effectively captures the model’s reliance on visual information versus language priors at the token level.

\label{sec:lb_exapmle}

\section{Attention Output Reorientation}
\label{sec:proof}
\begin{duplicate}    
Consider a layer $l$ within an LVLM, and let $h$ be the index of the attention head to be reinforced. Let $\widetilde{Z}_l$ be the input to the FFN module obtained with $\widetilde{A}_{l,h}=\alpha \cdot A_{l,h}$ 
 ($\alpha > 1$), $Z_l$ be the original input obtained with $A_{l,h}$, $Z_{l,h}$ be the pseudo-input obtained with only the $A_{l,h}$ component. Then it holds that $cos(\widetilde{Z}_l,Z_{l,h})>cos(Z_l, Z_{l,h})$. 
\end{duplicate}
\textit{Proof.} By partitioning the projection matrix $W_l^O$, we can further express the output of MHA as the sum of the contributions from each attention head:
\begin{equation}
\begin{aligned}
    &\text{MHA}_l(X_l) \\
    &\quad= [A_{l,1}(X_{l,1}), \cdots, A_{l,n_h}(X_{l,n_h})]W_l^O \\
    &\quad= A_{l,1}(X_{l,1})W_{l,1}^O+ \cdots+ A_{l,n_h}(X_{l,n_h})W_{l,h}^O.
\end{aligned}
\end{equation}
To simplify the notation , we define $\mathbf{x}$ and $\mathbf{y}$ as follows:  
\begin{equation}
\begin{aligned}
    \mathbf{x}&=\hat X_l+\text{MHA}_l(X_l), \\
    \mathbf{y}&=A_{l,h}(X_{l,h})W_{l,h}^O. \\
\end{aligned}
\end{equation}
We then substitute $\mathbf{x}$ and $\mathbf{y}$ for the variables in Equation \ref{eq:ffn} to derive $Z_l, \widetilde{Z}_l, Z_{l,h}$:
\begin{equation}
\begin{aligned}
    Z_l&=\hat g_t \frac{\mathbf{x}}{\|\mathbf{x}\|}, \\
    \widetilde{Z}_l&=\hat g_t \frac{\mathbf{x}+(\alpha-1)\mathbf{y}}{\|\mathbf{x}+(\alpha-1)\mathbf{y}\|}, \\
    Z_{l,h}&=\hat g_t \frac{\mathbf{y}}{\|\mathbf{y}\|}.
\end{aligned}
\end{equation}
Lastly, we prove that $cos(\widetilde{Z}_l,Z_{l,h})$ is greater than $cos(Z_l,Z_{l,h})$:

\begin{equation}
\begin{aligned}
    &cos(\widetilde{Z}_l,Z_{l,h})-cos(Z_l,Z_{l,h}) \\
    &=\frac{\langle \mathbf{x}+(\alpha-1)\mathbf{y},\mathbf{y}\rangle}{\|\mathbf{x}+(\alpha-1)\mathbf{y}\|\|\mathbf{y}\|} - 
           \frac{\langle \mathbf{x},\mathbf{y}\rangle}{\|\mathbf{x}\|\|\mathbf{y}\|} \\
    &=\frac{\langle \mathbf{x}+(\alpha-1)\mathbf{y},(\alpha-1)\mathbf{y}\rangle}{\|\mathbf{x}+(\alpha-1)\mathbf{y}\|\|(\alpha-1)\mathbf{y}\|} - 
           \frac{\langle \mathbf{x},(\alpha-1)\mathbf{y}\rangle}{\|\mathbf{x}\|\|(\alpha-1)\mathbf{y}\|} \\
    &=\frac{\langle \mathbf{x},\hat{\mathbf{y}}\rangle+\|\hat{\mathbf{y}}\|^2}{\|\mathbf{x}+\hat{\mathbf{y}}\|\|\hat{\mathbf{y}}\|} - 
           \frac{\langle \mathbf{x},\hat{\mathbf{y}}\rangle}{\|\mathbf{x}\|\|\hat{\mathbf{y}}\|} \\
    &>\frac{-\|\mathbf{x}\|+\|\hat{\mathbf{y}}\|}{\|\mathbf{x}+\hat{\mathbf{y}}\|} + 1 \\
    &=\frac{\|\mathbf{x}+\hat{\mathbf{y}}\|+\|-\hat{\mathbf{y}}\|-\|\mathbf{x}\|}{\|\mathbf{x}+\hat{\mathbf{y}}\|} \\
    &>0,
\end{aligned}
\end{equation}
which concludes the proof.

\section{Choice of Reinforced Layers}
\label{sec:layer_choice}
Since \citealp{chen2025image} and \citealp{zhang2024seeing} have highlighted the unique role of the second layer in integrating visual information through attention map analysis, we include this layer for VHR and further validated its significance in mitigating hallucinations. Table \ref{tab:ablate-layer} presents the results of ablation experiments conducted on this layer and the last few layers on LLaVA-1.5. The results show that enhancing layer1 and the deeper layers both significantly alleviate hallucinations, with the combination of both yielding even better results. This suggests that VHR in the shallow and deep layers alleviates hallucinations through distinct mechanisms. However, we note that applying VHR to the second layer does not universally improve performance; in some challenging benchmarks, its benefits diminish or even introduce instability. Further analysis of the layer-specific mechanisms in LVLMs remains an important direction for future work. 

\section{Choice of Scale Factor}
\label{sec:scale_factor}
Table \ref{tab:ablate-alpha} shows the results of ablation experiments on the scale factor $\alpha$ in VHR. When $\alpha$ is set to 2 or 3, hallucinations are effectively alleviated. However, as $\alpha$ increases to 4, excessive intervention disrupts the model’s behavior, causing anomalies in the hallucination metric. Conversely, when $\alpha < 1$, which weakens the contribution of attention heads sensitive to visual information, hallucinations become significantly more pronounced. This further confirms the crucial role of the attention heads identified based on the VHD scores in mitigating hallucinations in LVLMs. 

\begin{table}[t]
\centering
\resizebox{\linewidth}{!}{
\begin{tabular}{l|cc}
\toprule
 & CHAIR$_S \downarrow$ & CHAIR$_I \downarrow$ \\
\midrule
LLaVA-1.5 & 49.68 & 14.32 \\
w/ VHR on layer1 & 40.80 & 12.00 \\
w/ VHR on last 14 layers & 41.96 & 12.56 \\
w/ VHR on both & \textbf{33.32} & \textbf{9.71} \\
\bottomrule
\end{tabular}
}
\caption{Ablation study on the reinforced layers.}
\label{tab:ablate-layer}
\end{table}

\begin{table}[t]
\centering
\resizebox{\linewidth}{!}{
\begin{tabular}{l|ccc}
\toprule
 & CHAIR$_S\downarrow$ & CHAIR$_I\downarrow$ & Len \\
\midrule
LLaVA-1.5 & 49.68 & 14.32 & 83.06 \\
w/ VHR $\alpha=0.2$ & 63.28 & 21.04 & 86.42 \\
w/ VHR $\alpha=0.5$ & 55.80 & 17.32 & 84.47 \\
w/ VHR $\alpha=2$ & 33.32 & 9.71 & 81.33 \\
w/ VHR $\alpha=3$ & 27.04 & 8.68 & 88.31 \\
w/ VHR $\alpha=4$ & 3.64 & 2.01 & 144.54 \\
\bottomrule
\end{tabular}
}
\caption{Ablation study on the scale factor.}
\label{tab:ablate-alpha}
\end{table}

\begin{figure*}[t]
\centering
  \includegraphics[width=\linewidth]{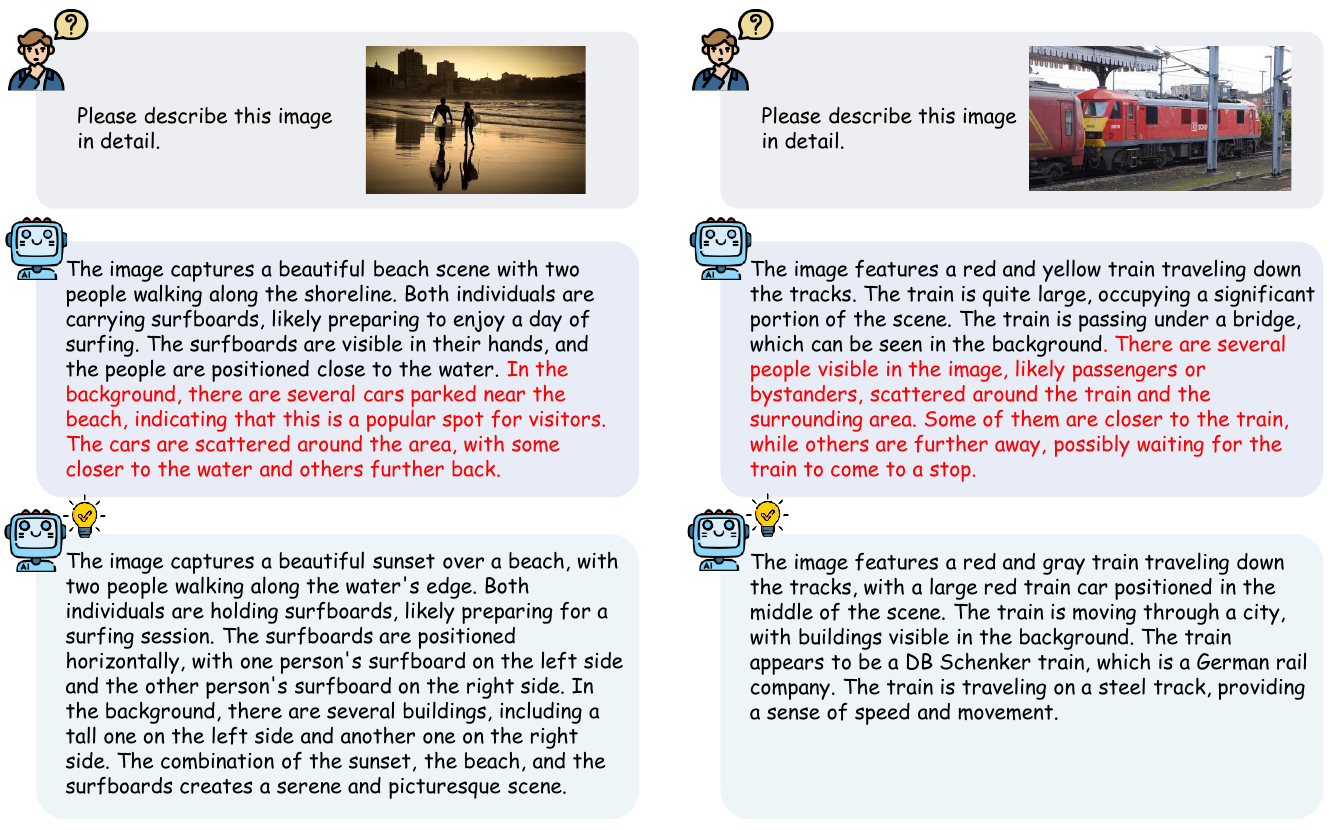}
  \caption{More examples of VHR eliminating the hallucinated content.}
  \label{fig:more_qualitative}
\end{figure*}

\section{Examples of VHD Scores}
\label{sec:vhd_examples}
\begin{figure*}[t]
\centering
  \includegraphics[width=\linewidth]{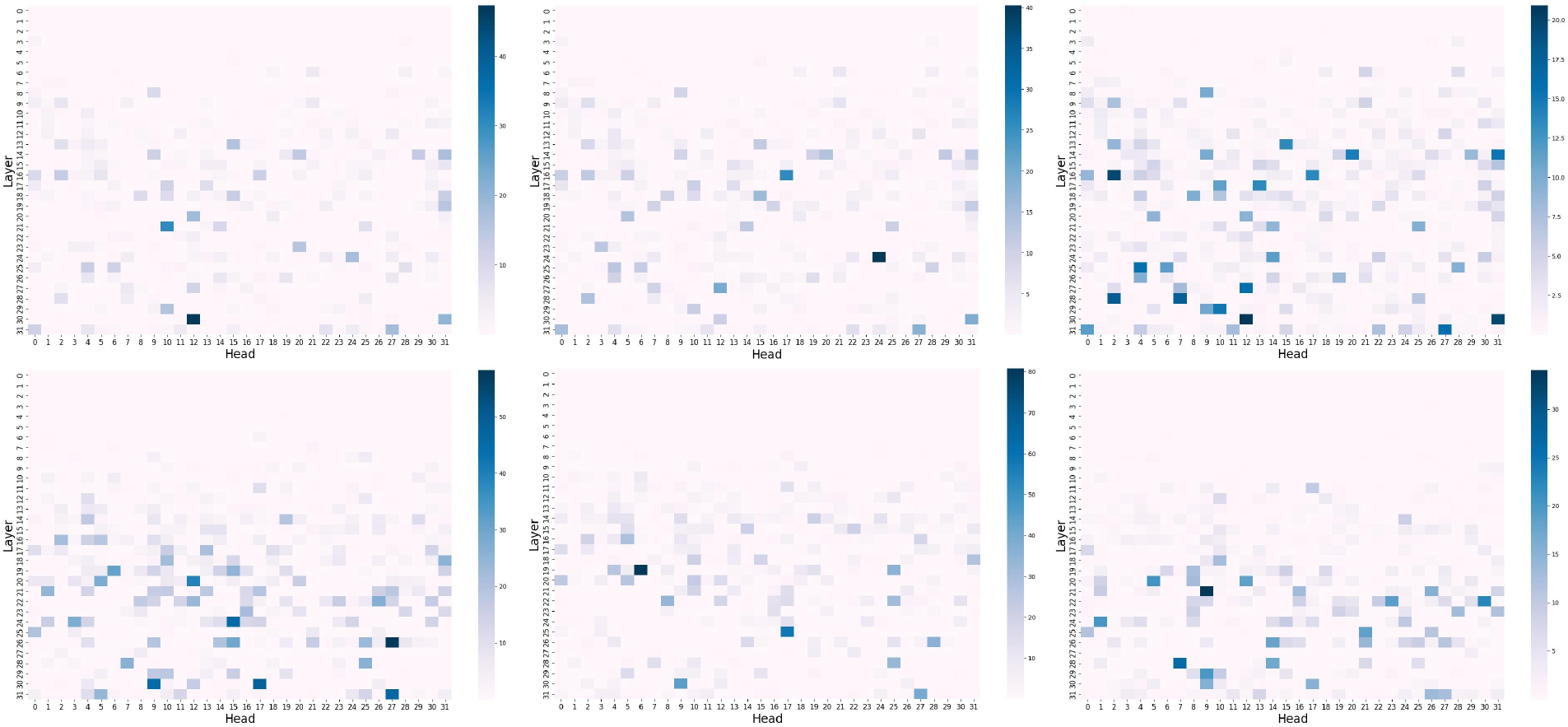}
  \caption{More examples of the VHD scores during different generation steps of different samples.}
  \label{fig:vhd_example}
\end{figure*}

As shown in Figure \ref{fig:vhd_example}, the first row presents the VHD scores during the first generation step across different samples, while the second row shows the VHD scores for different object terms generated within the same sample. It can be observed that VHD scores vary across different samples and generation steps; however, significant differences between the VHD scores of the attention heads within the model are consistently present.

\section{More Qualitative Results}
As shown in Figure \ref{fig:more_qualitative}, we present more examples that illustrate the effect of VHR in eliminating hallucinated objects. After incorporating VHR, the descriptions generated by the LVLMs faithfully align with the content of the images, while preserving the richness of the descriptions.

\section{Details on the GPT-4V Evaluation}

To evaluate the performance of LVLMs on LLaVA-Bench (In-the-Wild), we use GPT-4o as the evaluator. The prompt template adapted from \cite{gong2024damro} is shown in Table~\ref{tab:gpt-eval}, with an additional metric, \textit{Naturalness}, introduced to assess the fluency and coherence of the generated language. For each sample, GPT-4o is provided with the original image, the baseline LVLM output, and the output from the VHR-enhanced model. The evaluation focuses on three key aspects: accuracy, detailedness, and naturalness, with particular emphasis on the reduction of hallucinations in the VHR-enhanced responses compared to the baseline.

\begin{table*}[t]
\centering
\resizebox{\linewidth}{!}{
\begin{tabular}{l}
\hline
\textbf{GPT-4V Prompt}                                                                                                                                                                                                                                                                                                                                                                                                                                                                                                                                                                                               \\ \hline
\begin{tabular}[c]{@{}l@{}}You are required to score the performance of two AI assistants in describing a given image. You should pay\\extra attention to the hallucination, which refers to the part of descriptions that are inconsistent with the image\\content, such as claiming the existence of something not present in the image or describing incorrectly in terms\\of the counts, positions, or colors of objects in the image. Please rate the responses of the assistants on a scale of\\1 to 10, where a higher score indicates better performance, according to the following criteria:\end{tabular} \\
\begin{tabular}[c]{@{}l@{}}1: Accuracy: whether the response is accurate with respect to the image content. Responses with fewer\\hallucinations should be given higher scores.\end{tabular}                                                                                                                                             \\
\begin{tabular}[c]{@{}l@{}}2: Detailedness: whether the response is rich in necessary details. Note that hallucinated descriptions should\\not count as necessary details.\end{tabular}                                                                                                                                                                                                                    \\
\begin{tabular}[c]{@{}l@{}}3: Naturalness: assess the language quality, focusing on: fluency of sentence structure, appropriateness of word\\choice, smoothness of language flow, absence of awkward or unnatural phrasing.\end{tabular}                                                                                                  \\

\begin{tabular}[c]{@{}l@{}}Please output the scores for each criterion, containing only two values indicating the scores for Assistant 1 and\\2, respectively. The two scores are separated by a space. Following the scores, please provide an explanation of\\your evaluation, avoiding any potential bias and ensuring that the order in which the responses were presented\\does not affect your judgment.\end{tabular}                                                                         \\
{[}Assistant 1{]}                                                                                                                                                                                                                                                                          \\
\{\}                                                                                                                                                  \\
{[}End of Assistant 1{]}                                                                                                                                                                                             \\
       \\
{[}Assistant 2{]}                                                                                                                                                                                                                                                                   \\
\{\}                                                                                                                                                                                                                 \\
{[}End of Assistant 2{]}                                                                                                                                                                                                  \\
   \\
Output format:                                                                                                                                                                                                                     \\
Accuracy:                                                                                                                                                                                                                \\
Reason:                                                                                                                                                                                                                   \\
Detailedness:                                                                                                                                           \\
Reason:                                                                                                                                                                                                                   \\
Naturalness:                                                                                                                                           \\
Reason:                                                                                                                                                                                                                   \\ \hline
\end{tabular}
}
\caption{The prompt used for GPT-4V evaluation.}
\label{tab:gpt-eval}
\end{table*}

\end{document}